\documentclass[conference]{IEEEtran}
\ifCLASSINFOpdf
\else
\fi
\usepackage{times}

\usepackage{multicol}
\usepackage[bookmarks=true]{hyperref}
\usepackage[pdftex]{graphicx}
\usepackage[ruled,vlined]{algorithm2e}
\usepackage[bookmarks=true]{hyperref}
\usepackage{amsmath}
\usepackage{siunitx}
\usepackage[caption=false, font=footnotesize]{subfig}
\usepackage[dvipsnames]{xcolor}
\usepackage{float}
\begin{document}
\bstctlcite{IEEEexample:BSTcontrol}
\IEEEoverridecommandlockouts
\title{Sniffy Bug: A Fully\\ Autonomous Swarm of Gas-Seeking\\ Nano Quadcopters in Cluttered Environments}
\author{Bardienus P. Duisterhof$^{1}$~~~Shushuai Li$^{1}$~~~Javier Burgués$^{2}$~~~Vijay Janapa Reddi$^{3}$~~~Guido C.H.E. de Croon$^{1}$
\thanks{$^{1}$ Delft University of Technology, $^{2}$ University of Barcelona, $^{3}$ Harvard University. Email: \href{b.p.duisterhof@gmail.com}{b.p.duisterhof@gmail.com} or \href{g.c.h.e.decroon@tudelft.nl}{g.c.h.e.decroon@tudelft.nl}}
}



%

\maketitle

\begin{abstract}
Nano quadcopters are ideal for gas source localization (GSL) as they are safe, agile and inexpensive. However, their extremely restricted sensors and computational resources make GSL a daunting challenge. In this work, we propose a novel bug algorithm named `Sniffy Bug', which allows a fully autonomous swarm of gas-seeking nano quadcopters to localize a gas source in an unknown, cluttered and GPS-denied environments. The computationally efficient, mapless algorithm foresees in the avoidance of obstacles and other swarm members, while pursuing desired waypoints. The waypoints are first set for exploration, and, when a single swarm member has sensed the gas, by a particle swarm optimization-based procedure. We evolve all the parameters of the bug (and PSO) algorithm, using our novel simulation pipeline, `AutoGDM'. It builds on and expands open source tools in order to enable fully automated end-to-end environment generation and gas dispersion modeling, allowing for learning in simulation. Flight tests show that Sniffy Bug with evolved parameters outperforms manually selected parameters in cluttered, real-world environments. Videos: \url{https://bit.ly/37MmtdL}
\end{abstract} 


\IEEEpeerreviewmaketitle

\section{Introduction}

Gas source localization (GSL) by autonomous robots is important for search and rescue and inspection, as it is a very dangerous and time-consuming task for humans. A swarm of nano quadcopters is an ideal candidate for GSL in large, cluttered, indoor environments. The quadcopters' tiny size allows them to fly in narrow spaces, while operating as a swarm enables them to spread out and find the gas source much quicker than a single robot would.


To enable a fully autonomous gas-seeking swarm, the nano quadcopters need to navigate in unknown, cluttered, GPS-denied environments by avoiding obstacles and each other. Currently, indoor swarm navigation is still challenging and an active research topic even for larger quadcopters ($>$ 500 grams)~\cite{10.3389/frobt.2020.00018,Xu_2020}. State-of-the-art methods use heavy sensors like LiDAR and high-resolution cameras to construct a detailed metric map of the environment, while also estimating the robot's position for navigation with Simultaneous Localization And Mapping (SLAM, e.g., ORB-SLAM2~\cite{7946260}). These methods do not fit within the extreme resource restrictions of nano quadcopters. The payload of nano quadcopters is in the order of grams, ruling out heavy, power-hungry sensors like LiDAR. Furthermore, SLAM algorithms typically require GBs of memory~\cite{8460558} and need considerable processing power. One of the most efficient implementations of SLAM runs real-time (18.21 fps) on an ODROID-XU4~\cite{8875145}, which has a CPU with a 4-core @ 2GHz plus a 4-core @1.3GHz. These properties rule out the use of SLAM on nano quadcopters such as the BitCraze Crazyflie, which has an STM32F405 processor with 1MB of flash memory and a single core @168MHz. As a result of the severe resource constraints, previous work has explored alternative navigation strategies. A promising solution was introduced in~\cite{McGuireeaaw9710}, in which a bug algorithm enabled a swarm of nano quadcopters to explore unknown, cluttered environments and come back to the starting location.



\begin{figure}[t]
    \vspace{2.2mm}
    \centering
    \includegraphics[width=0.9\linewidth]{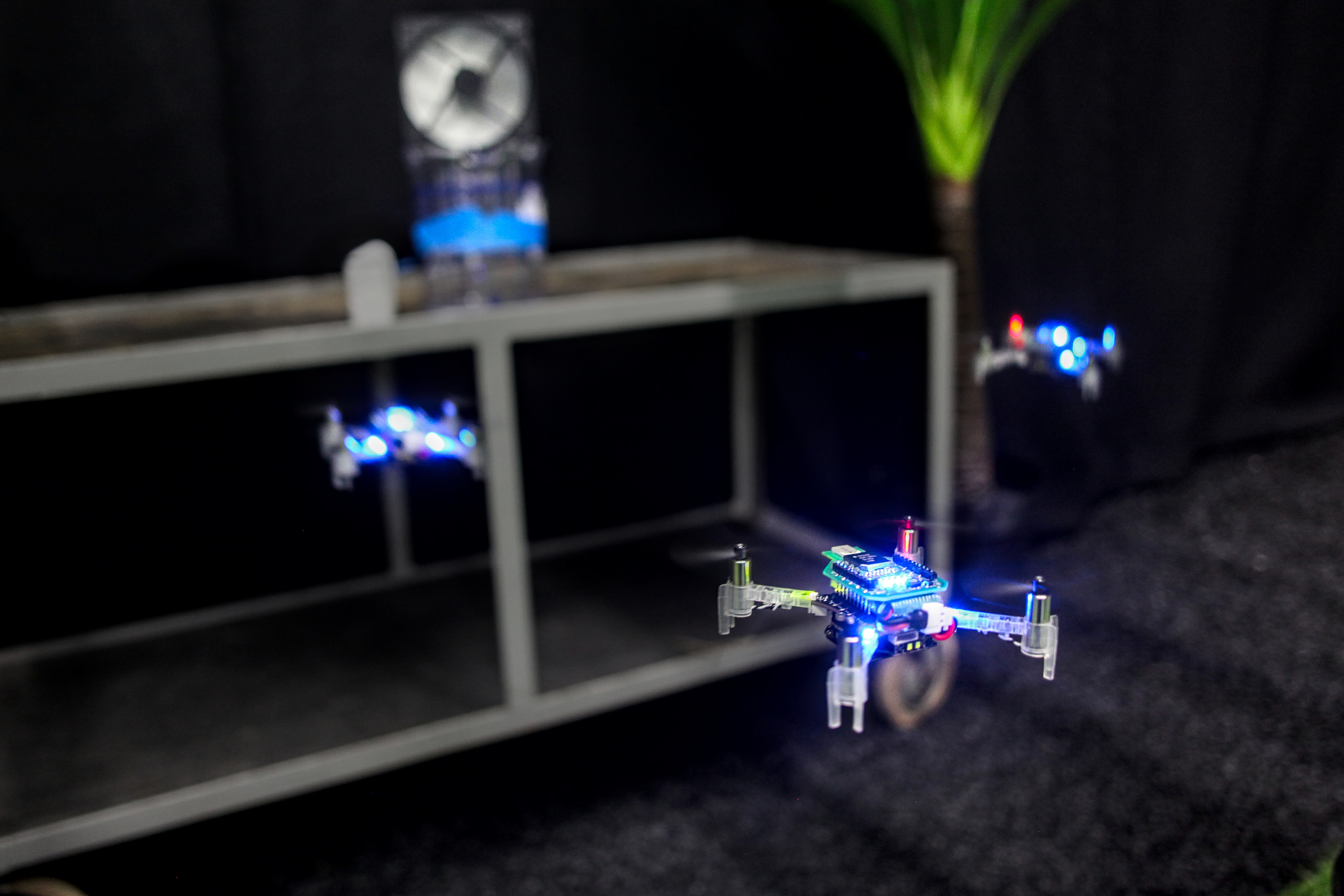}
        \caption{A fully autonomous and collaborative swarm of gas-seeking nano quadcopters, finding and locating an isopropyl alcohol source. The source is visible in the background: a spinning fan above a can of isopropyl alcohol. }
    \label{fig:swarm}
    \vspace*{-0.2cm}
\end{figure}





Besides navigating, the swarm also needs a robust strategy to locate the gas source, which by itself is a highly challenging task. This is mostly due to the complex spreading of gas in cluttered environments. Moreover, current sensors have poor quality compared to animals' smelling capabilities~\cite{10.1088/1748-3190/abbd81}, which is further complicated by the propellers' own down-wash \cite{ercolani20203d}.

Various solutions to odor source localization have been studied. Probabilistic GSL strategies \cite{Song2020,Vergassola2007} usually internally maintain a map with the probabilities for the odor source location and often simulate the gas distribution. This is computationally challenging for nano quadcopters, a situation that deteriorates when they have to operate in environments with complex shapes, obstacles and a complex airflow field. In contrast, bio-inspired finite-state machines~\cite{KUWANA1999195,10.2307/24950326} have very low computational requirements, though until now they have focused on deploying a single agent~\cite{Lochmatter2009}. Moreover, reinforcement and evolutionary learning approaches have been investigated, mostly in simulation~\cite{Beer1992,Izquierdo2008,Izquierdo2010,DeCroon2013,Wei2012}, but a few works also transferred the learned policy to obstacle-free environments in the real world \cite{Kuwana1996,hayes_martinoli_goodman_2003}. A limiting factor for approaches that learn in simulation is that gas dispersion modeling has been a time-intensive process, requiring domain knowledge for accurate modeling. Only few environments have been made available to the public~\cite{monroy_2017_gaden,plumesim}, whereas learning algorithms require many different environments. 

Due to the difficulty of the problem in the real world, most often the experiments involve a single robot seeking for an odor source in an obstacle-free environment of modest size, e.g., in the order of 4 $\times$ 4 $m$~\cite{CHEN2020107482,Asenov2019,Moraud2010,10.1371/journal.pcbi.1003861}. Few experiments have been performed in larger areas involving multiple robots. Very promising in this area is the use of particle swarm optimization (PSO)~\cite{4168420}, as it is able to deal with the local maxima in gas concentration that arise in more complex environments. Closest to our work is an implementation of PSO on a group of large (\SI{2.95}{\kilo\gram}), outdoor flying quadcopters~\cite{Steiner2019} , using LiDAR and GPS for navigation. 

In this article, we introduce a novel PSO-powered bug algorithm, \emph{Sniffy Bug}, to tackle the problem of GSL in challenging, cluttered, and GPS-denied environments. The nano quadcopters execute PSO by estimating their relative positions and by communicating observed gas concentrations to each other using onboard Ultra-Wideband (UWB) \cite{li2020autonomous}. In order to optimize the parameters of Sniffy Bug with an artificial evolution, we also develop and present the first fully Automated end-to-end environment generation and Gas Dispersion Modeling pipeline, which we term \emph{AutoGDM}. We validate our approach with robot experiments in cluttered environments, showing that evolved parameters outperform manually tuned parameters. This leads to the following three contributions:
\begin{enumerate}
    \item The first robotic demonstration of a swarm of autonomous nano quadcopters locating a gas source in unknown, GPS-denied environments with obstacles.
    \item A novel, computationally highly efficient bug algorithm, Sniffy Bug, of which the parameters are evolved for PSO-based gas source localization in unknown, cluttered and GPS-denied environments.
    \item  The first fully automated environment generation and gas dispersion modeling pipeline, AutoGDM.
\end{enumerate}

In the remainder of this article, we explain our methodology (Section~\ref{sec:method}), present simulation and flight results (Section~\ref{sec:results}), and draw conclusions (Section~\ref{sec:conclusion}).

\section{Method}
\label{sec:method}

\subsection{System Design}
\label{sec:system_design}
\begin{figure}
    \centering
    \includegraphics[width=0.8\linewidth]{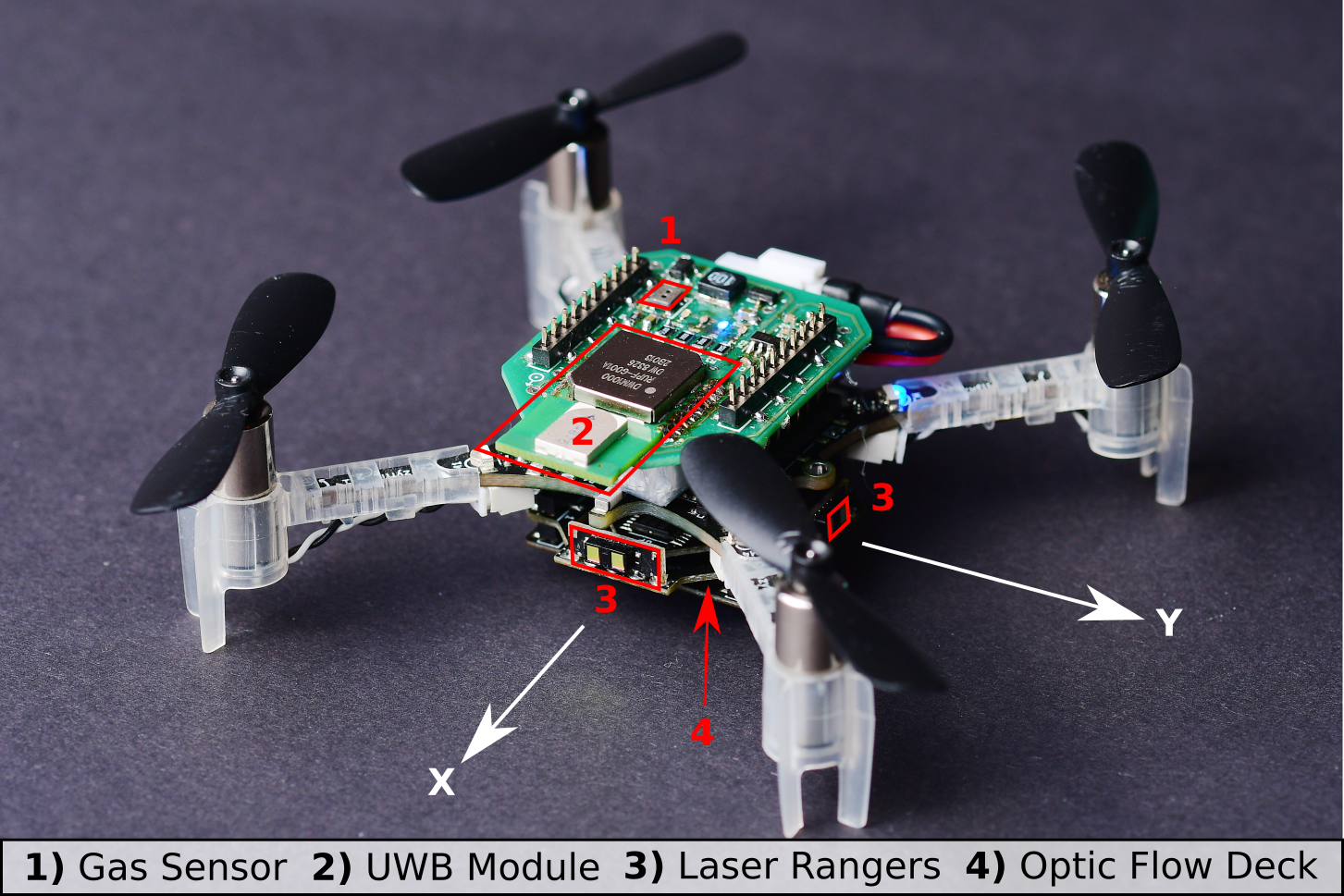}
    \caption{A \SI{37.5}\gram~nano quadcopter, capable of autonomous waypoint tracking, obstacle avoidance, relative localization, communication and gas sensing. }
    \label{fig:system_design}
     \vspace*{-0.5cm}
\end{figure}

Our \SI{37.5}\gram~Bitcraze CrazyFlie~\cite{8046794} nano quadcopter (Figure~\ref{fig:system_design}), is equipped with sensors for waypoint tracking, obstacle avoidance, relative localization, communication and gas sensing. We use the optical flow deck and IMU sensors for estimating the drone's state and tracking waypoints. Additionally, we use the BitCraze multiranger deck with four laser range sensors in the drone's positive and negative $x$ and $y$ axis (Figure~\ref{fig:system_design}), to sense and avoid obstacles. Finally, we have designed a custom, open-source PCB, capable of gas sensing and relative localization. It features a Figaro TGS8100 MEMS MOX gas sensor, which is lightweight, inexpensive, low-power, and was previously deployed onboard a nano quadcopter~\cite{Burgues2019}. We use it to seek isopropyl alcohol, but it is sensitive to many other substances, such as carbon monoxide. The TGS8100 changes resistance based on exposure to gas, which can be computed according to Equation~\ref{eq:gas_sensor}. 
\begin{equation}
    R_{s} = \left( \frac{V_{c}}{V_{RL}}-1 \right) \cdot R_{L}
\label{eq:gas_sensor}
\end{equation}
Here $R_{s}$ is the sensor resistance, $V_{c}$ circuit voltage (\SI{3.0}\volt), $V_{RL}$ the voltage drop over the load resistor in a voltage divider, and $R_{L}$ is the load resistor's resistance (\SI{68}{\kilo\ohm}). Since different sensors can have different offsets in the sensor reading, we have designed our algorithm not to need absolute measurements like a concentration in ppm. From now on, we report a corrected version of $V_{RL}$, where higher means more gas. $V_{RL}$ is corrected by its initial low-passed reading, without any gas present, in order to correct sensor-specific bias. 

For relative ranging, communication, and localization, we use a Decawave DWM1000 UWB module. Following~\cite{li2020autonomous}, an extended Kalman filter (EKF) uses onboard sensing from all agents, measuring velocity, yaw rate, and height, which is fused with the UWB range measurements. It does not rely on external systems or magnetometers. Additionally, all agents are programmed to maintain constant yaw, as it further improves the stability and accuracy of the estimates. 

\subsection{Algorithm Design}
\label{sec:algo_design}
We design a highly efficient algorithm, both from a computational and sensor perspective. We generate waypoints in the reference frame of each agent using PSO, based on the relative positions and gas readings of all agents. The reference frame of the agent is initialized just before takeoff, and is maintained using dead reckoning by fusing data from the IMU and optic flow sensors. While the reference frame does drift over time, only the drift since the last seen `best waypoint' in PSO will be relevant, as it will be saved in the drifted frame.

The waypoints are tracked using a bug algorithm that follows walls and other objects, moving safely around them. Each agent computes a new waypoint if it reaches within $d_{wp}$ of its previous goal, if the last update was more than $t_{wp}$ seconds ago, or when one of the agents smells a concentration superior to what has been seen by any agent during the run, and higher than the pre-defined detection threshold. A detection threshold is in place to avoid reacting based on sensor drift and noise. A timeout time, $t_{wp}$, is necessary, as the predicted waypoint may be unreachable (e.g., inside a closed room). We term each new waypoint, generated if one of the above criteria is met, a `waypoint iteration' (e.g., waypoint iteration five is the fifth waypoint generated during that run).

\subsubsection{Waypoint generation}
Each agent computes its next waypoint according to Equation~\ref{eq:goal}.
\begin{equation}
   \mathbf{g}_{i,j} = \mathbf{x}_{i,j} + \mathbf{v}_{i,j}
   \label{eq:goal}
\end{equation}
Here $\mathbf{g}_{i,j}$ is the goal waypoint of agent $i$, in iteration $j$, $\mathbf{x}_{i,j}$ its position when generating the waypoint, and $\mathbf{v}_{i,j}$ its `velocity vector'. The velocity vector is determined depending on the drone's mode, which can be either `exploring', or `seeking'. `Exploring' is activated when none of the agents has smelled gas, while `seeking' is activated when an agent has detected gas. During exploration, $\mathbf{v}_{i,j}$ is computed with Equation~\ref{eq:no_gas}:
\begin{equation}
    \mathbf{v}_{i,j} =\omega' (\mathbf{g}_{i,j-1} - \mathbf{x}_{i,j})+ r_{r}\left(\mathbf{r}_{i,j} - \mathbf{x}_{i,j} \right)
    \label{eq:no_gas}
\end{equation}
Here $\mathbf{g}_{i,j-1}$ is the goal computed in the previous iteration and $\mathbf{r}_{i,j}$ a random point within a square of size $r_{rand}$ around the agent's current position. A new random point is generated each iteration. Finally, $\omega'$ and $r_r$ are scalars that impact the behavior of the agent. $\mathbf{g}_{i,j}$ and $\mathbf{v}_{i,j}$ are initialized randomly in a square of size $r_{rand}$ around the agent. Intuitively, Equation~\ref{eq:no_gas} shows the new velocity vector is a weighted sum of: 1) a vector toward the previously computed goal (also referred to as inertia), and 2) a vector towards a random point.
\begin{figure*}
    \centering
    \includegraphics[height=\linewidth,angle=270]{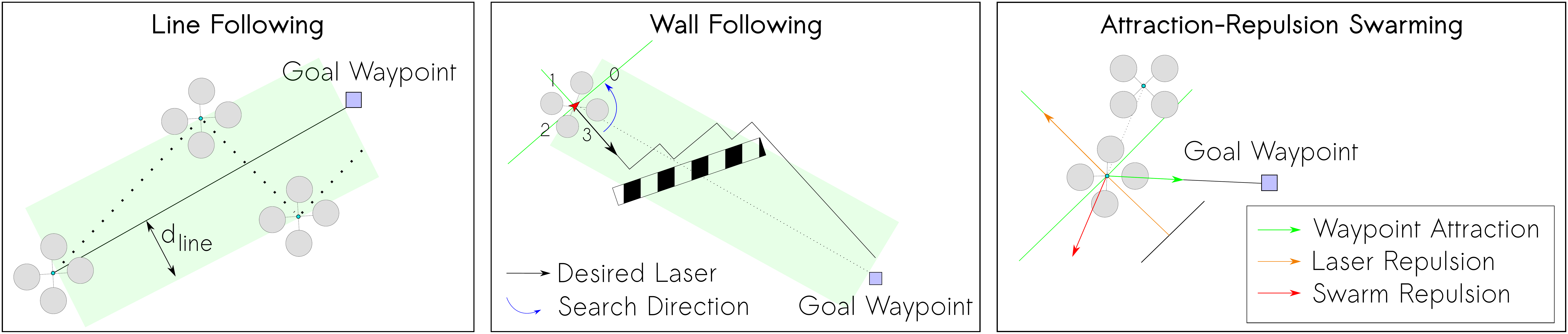}
    \caption{Sniffy Bug's three states: line following, wall following, and attraction-repulsion swarming.}
    \label{fig:sniffy_bug}
\end{figure*}

After smelling gas, i.e., one of the agents detects a concentration above a pre-defined threshold, we update the waypoints according to Equation~\ref{eq:gas}.
\begin{equation}
\begin{split}
    \mathbf{v}_{i,j} = \omega (\mathbf{g}_{i,j-1} - \mathbf{x}_{i,y})+\varphi_{p}  \alpha_{i,j}\left(\mathbf{p}_{i,j}-\mathbf{x}_{i,j}\right)+ \\ \varphi_{g}   \beta_{i,j}\left(\mathbf{s}_{j}-\mathbf{x}_{i,j}\right)
\end{split}
\label{eq:gas}
\end{equation}
Here $\mathbf{p}_{i,j}$ is the waypoint at which agent $i$ has seen its highest concentration so far, up to iteration $j$. $\mathbf{s}_{j}$ is the swarm's best seen waypoint, up to iteration $j$. $\alpha_{i,j}$ and $\beta_{i,j}$ are random values between 0 and 1, generated for each waypoint iteration for each agent. Finally, $\varphi_p$ and $\varphi_g$ are scalars that impact the behavior of the agent. So again, more intuitively, the vector towards the next waypoint is a weighted sum of the vectors towards its previously computed waypoint, the best seen position by the agent, and the best seen position by the swarm. As we will see later, this allows the swarm to converge to high concentrations of gas, whilst still exploring.

\subsubsection{Waypoint tracking}
Tracking waypoints in cluttered environments is hard due to the limited sensing and computational resources. Sniffy Bug is designed to operate at constant yaw, and consists of three states (Figure~\ref{fig:sniffy_bug}): 1) Line Following, 2) Wall Following, and 3) Attraction-Repulsion Swarming.

\textbf{Line Following} -- When no obstacles are present, the agent follows a virtual line towards the goal waypoint, making sure to only move in directions where it can `see' using a laser ranger, improving safety. The agent makes sure to stay within a distance $d_{line}$ from the line, moving as shown in Figure~\ref{fig:sniffy_bug}.

\textbf{Wall Following} -- When the agent detects an obstacle, and no other agents are close, it will follow the object similar to other bug algorithms~\cite{MCGUIRE2019103261}. Sniffy Bug's wall-following stage is visible in Figure~\ref{fig:sniffy_bug}. It is terminated if one of the following criteria is met: 1) a new waypoint has been generated, 2) the agent has avoided the obstacle, or 3) another agent is close. In case 1 and 2 line following is selected, whereas in case 3 attraction-repulsion swarming is activated. Figure~\ref{fig:sniffy_bug} illustrates wall following in Sniffy Bug. The agent starts by computing $desired laser$, which is the laser direction that points most directly to the goal waypoint, laser 3 in this case. It then determines the initial search direction in the direction of the waypoint, anti-clockwise in this case. The agent now starts looking for laser rangers detecting a safe value (above $d_{laser}$), starting from lasers 3, in the anti-clockwise direction. As a result, we follow the wall in a chainsaw pattern, alternating between lasers 3 and 0. Next, the agent uses odometry to detect that it has avoided the obstacle, by exiting and re-entering the green zone, while getting closer to the goal waypoint.


\textbf{Attraction-Repulsion Swarming} -- When the agent detects at least one other agent within a distance $d_{swarm}$, it uses attraction-repulsion swarming to avoid other agents and objects, while tracking its goal waypoint. This state is terminated when no agent is within $d_{swarm}$, selecting `line following' instead. As can be seen in Figure~\ref{fig:sniffy_bug} and Equations~\ref{eq:AR},\ref{eq:v_command}, the final commanded velocity vector is a sum of repulsive forces away from low laser readings and close agents, while exerting an attractive force to the goal waypoint. 

\begin{scriptsize}
  \begin{equation}
     \begin{split}
    \mathbf{A}_{i,t} =  \sum_{k=0}^{\#agents}k_{swarm}\cdot  Relu\left(d_{swarm}-\Vert\mathbf{x}_{i,k,t} \Vert  \right)\cdot  \frac{\mathbf{x}_{i,k,t}}{\Vert\mathbf{x}_{i,k,t} \Vert } \\ + 
    \sum_{k=0}^{3} k_{laser}\cdot Relu \left(d_{laser}'-l_{k,t}  \right)\cdot R\left(\frac{k+2}{2}\pi\right)\cdot\mathbf{i}\\
    +
    \frac{\mathbf{g}_{i,t}}{\Vert \mathbf{g}_{i,t} \Vert}\cdot V_{desired}
    \end{split}
    \label{eq:AR}
  \end{equation}  
\end{scriptsize}
In Equation~\ref{eq:AR}, $\mathbf{A}_{i,t}$ is the attraction vector of agent $i$ at time step (so not iteration) $t$, specifying the motion direction. Each time step the agent receives new estimates and re-computes $\mathbf{A}_{i,t}$. The first term results in repulsion away from other agents that are closer than $d_{swarm}$, while the second term adds repulsion from laser rangers seeing a value lower than $d'_{laser}$, and the third term adds attraction to the goal waypoint. 

In the first term, $k_{swarm}$ is the swarm repulsion gain, and $d_{swarm}$ is the threshold to start avoiding agents. $\mathbf{x}_{i,k,t} $ is the vector between agent $i$ and agent $k$, at time step $t$. The rectified linear unit ($Relu$) makes sure only close agents are repulsed. In the second term, $k_{laser}$ is the laser repulsion gain, and $d_{laser}'$ is the threshold to start repulsing a laser. $l_{k,t}$ is the laser reading $k$ at step $t$, numbered according to Figure~\ref{fig:sniffy_bug}. $Relu$ makes sure only lasers recording values lower than $d_{laser}'$ are repulsed. $R(\cdot)$ is the rotation matrix, used to rotate $\mathbf{i}$ in the direction away from laser $k$, such that the second term adds repulsion away from low lasers. The third term adds attraction from the agent's current position to the goal. $\mathbf{g}_{i,t}$ is the vector from agent $i$ to the goal waypoint, at time step $t$. This term is scaled to be of size $V_{desired}$, which is the desired velocity, a user-defined scalar. As a last step, we normalize $\mathbf{A}_{i,t}$ to have size $V_{desired}$ too, using Equation~\ref{eq:v_command}. 
  \begin{equation}
      \mathbf{V}_{command} = \frac{\mathbf{A}_{it}}{\Vert{\mathbf{A}_{it}\Vert}}\cdot V_{desired}
      \label{eq:v_command}
  \end{equation}
 
Here $\mathbf{V}_{command}$ is the velocity vector sent to the low-level control loops. Commanding a constant speed prevents both deadlocks in which the drones hover in place and peaks in velocity that induce collisions.


\subsection{AutoGDM}
\label{sec:autoGDM}
\begin{figure}
    \centering
    \includegraphics[height=3.5in,angle=270]{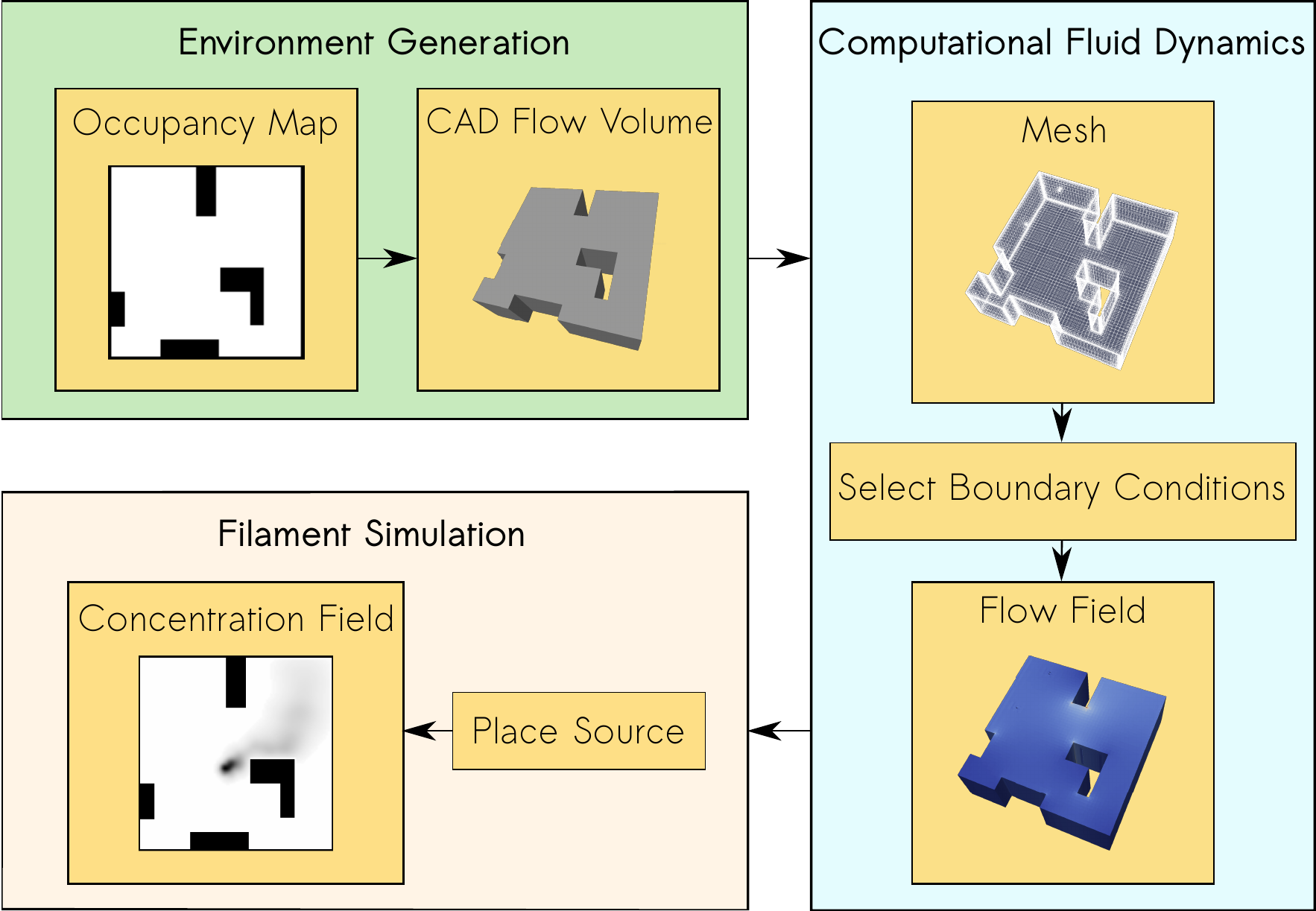}
    \caption{AutoGDM, a fully automated environment generation and gas dispersion modeling pipeline.}
    \label{fig:AutoGDM}
    \vspace{-4mm}
\end{figure}

Fully automated gas dispersion modeling based on Computational Fluid Dynamics (CFD) requires three main steps: 1) Environment generation, 2) CFD, and 3) Filament simulation (Figure~\ref{fig:AutoGDM}). 
\subsubsection{Environment Generation}
We use the procedural environment generator described in~\cite{MCGUIRE2019103261}, which can generate environments of any desired size with a number of requested rooms. Additionally, AutoGDM allows users to insert their own 2D binary occupancy images, making it possible to model any desired 2.5D environment by extruding the 2D image. 

\subsubsection{CFD}
CFD consists of two main stages, i.e., meshing and solving (flow field generation). We use the open-source package OpenFOAM~\cite{JASAK200989} for both stages. To feed the generated environments into OpenFOAM, the binary occupancy maps are converted into 3D triangulated surface geometries of the flow volume. We detect the largest volume in the image and declare it as our flow volume and test area. We create a mesh using OpenFOAM \texttt{blockMesh} and \texttt{snappyHexMesh}, and assign inlet and outlet boundary conditions randomly to vertical surfaces. Finally, we use OpenFOAM \texttt{pimpleFOAM} to solve for kinematic pressure, $p$, and the velocity vector, $U$.

\subsubsection{Filament simulation}
In the final stage of AutoGDM, we use the GADEN ROS package~\cite{monroy_2017_gaden} to model a gas source based on filament dispersion theory. It releases filaments that expand over time, and disappear when they find their way to the outlet. The expansion of the filaments and the dispersion rate of the source (i.e., filaments dispersed per second), is random within a user-defined range.

\subsection{Evolutionary Optimization}
\label{sec:evolution_setup}
We feed the generated gas data into Swarmulator\footnote{\href{https://github.com/coppolam/swarmulator}{https://github.com/coppolam/swarmulator}}, an open source lightweight C++ simulator for simulating swarms. The agent is modelled as a point mass, which is velocity-controlled using a P controller. We describe both the environment and laser rays as a set of lines, making it extremely fast to model laser rangers. An agent `crashes' when one of its laser rangers reads less than \SI{0.1}\meter~or when another agent is closer than \SI{0.5}\meter. The agents are fed with gas data directly from AutoGDM, which is updated every \SI{1.0}\second~in simulation time.

Using this simulation environment, we evolve the parameters of Sniffy Bug with the `simple genetic algorithm' from the open-source PyGMO/PAGMO package~\cite{pygmo}. The population consists of 50 individuals and is evolved for 400 generations. The algorithm uses tournament selection, mating through exponential crossover and mutation using a polynomial mutation operator. The mutation probability is 0.1, while the crossover probability is 0.9. The genome consists of 13 different variables, as shown in Table~\ref{tab:evo_vars}, including their ranges set during evolution. Parameters that have a physical meaning when negative are allowed to be negative, while variables such as time and distance need to be positive. 

Each agent's cost is defined as its average distance to source with an added penalty (+ 1.0) for a crash. Even without a penalty the agents will learn to avoid obstacles to some capacity, but a penalty stimulates prudence. Other metrics like `time to source' were also considered, but we found average distance to work best and to be most objective. It leads to finding the most direct paths and staying close to the source. 

In each generation, we select $n$ environments out of the total of $m$ environments generated using AutoGDM. As considerable heterogeneity exists between environments, we may end up with a controller that is only able to solve easy environments. This is known as the problem of hard instances. To tackle this problem, we study the effect of `doping'~\cite{doping} on performance in simulation. When using doping, the probability of  selecting environment number $i$ is: 
\begin{equation}
    P(i) = \frac{D(i)}{\sum_{k=0}^{m} D(k)}
\label{eq:doping}
\end{equation}

$D(i)$ is the `difficulty' of environment $i$, computed based on previous experience. If environment $i$ is selected to be evaluated, we save the median of all 50 costs in the population. We use median instead of mean to avoid a small number of poor-performing agents to have a large impact. $D(i)$ is the mean of the last three recorded medians. If no history is present, $D(i)$ is the average of all difficulties of all environments. Using Equation~\ref{eq:doping} implies that we start with an even distribution, but over time give environments with greater difficulty a higher chance to be selected in each draw. When not using doping, we set $P(i) = \frac{1}{m}$, resulting in a uniform distribution. 

\begin{figure}
    \centering
    \includegraphics[width=\linewidth]{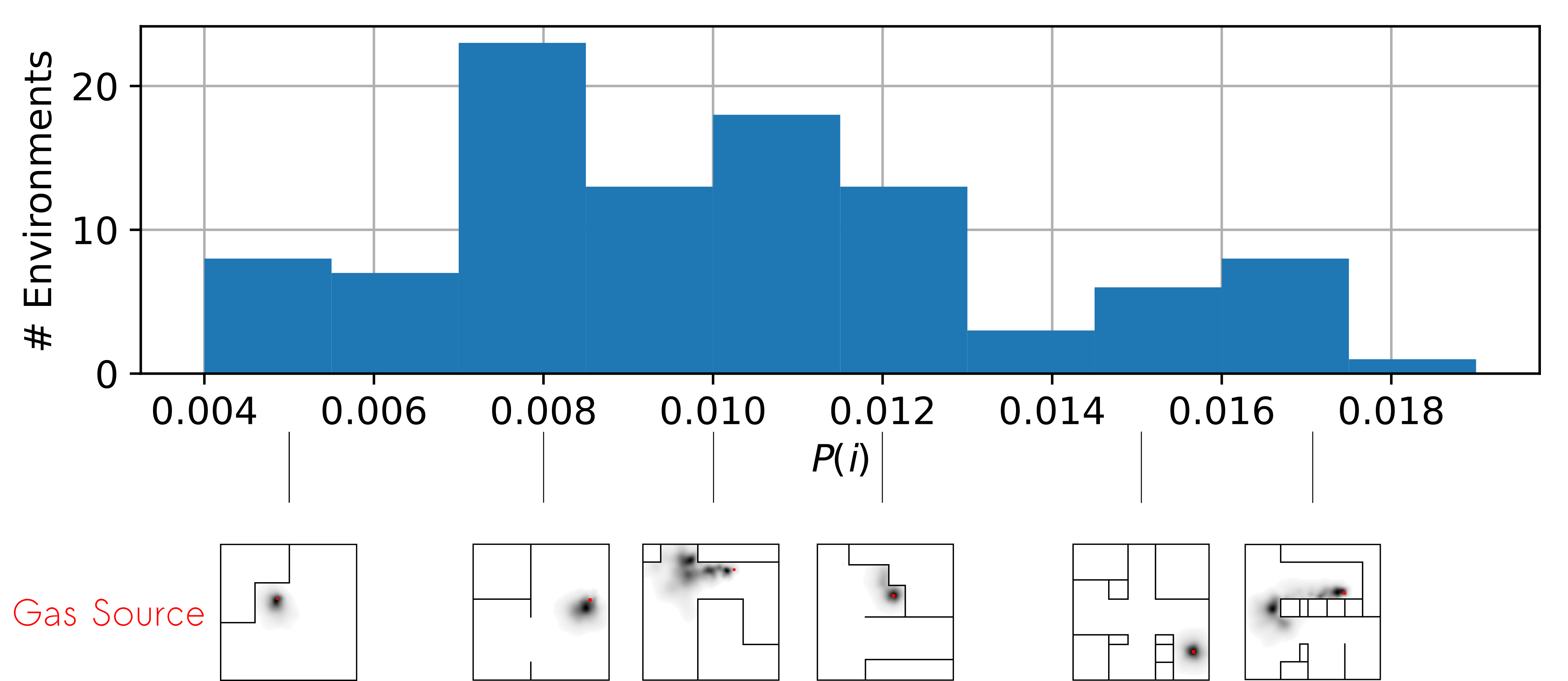}
    \caption{Environment selection probability $P(i)$ for all 100 environments at the end of evolution. The environments below the x-axis show that harder environments, with higher $P(i)$, contain more obstacles and local optima. }
    \label{fig:doping_props}
    \vspace{-3mm}
\end{figure}

\section{Results}
\label{sec:results}
\begin{table}[]
\begin{tabular}{l|l|l|l}
 Variable & Manually Selected   & Evolved  & Evolution range \\ \hline
$\omega$ & 0.5 & 0.271  & [-5,5]  \\
$\varphi_p$ & 0.8  &-0.333  & [-5,5] \\
$\varphi_g$ &  2.0 &1.856 &[-5,5]  \\
$\omega'$ &  0.3 & 1.571 &[-5,5]  \\
$r_r$ &  0.7 & 2.034 & [0,5]  \\
$t_{wp}$ & 10.0  & 51.979  & [0,100]  \\
$d_{wp}$  & 0.5  & 2.690 & [0,5] \\
$d_{laser}$  & 1.5  & 1.407 & [0,5] \\
$d_{swarm}$  &  1.5 &0.782  & [0,5]  \\
$d_{line}$  &  0.2 &0.469  & [0,1]  \\
$k_{laser}$  &  5.0 & 16.167 & [0,20] \\
$k_{swarm}$  &  15.0 &10.032  & [0,20] \\
$d_{laser}'$  &  1.5 & 0.594 & [0,5]
\end{tabular}
\vspace{2mm}
\caption{Parameters evolved in evolution using doping, consult Section~\ref{sec:algo_design} for the meaning of the variables.}
\label{tab:evo_vars}
 \vspace*{-0.6cm}
\end{table}
\subsection{Training in Simulation}
\label{sec:training_sim}
For evolution, we used AutoGDM to randomly generate 100 environments of 10 $\times$ 10 \si{\meter}~in size, the size of our experimental arena. This size is sufficiently large for creating environments with separate rooms, in which local maxima of gas concentration may exist. At the same time, it is sufficiently small for exploration by a limited number of robots. We use 3 agents, with $V_{desired} =$  \SI{0.5}{\meter/\second}. Figure~\ref{fig:doping_props} shows that the generated environments differ in obstacle configuration and gas distribution. During each generation, every genome is evaluated on a random set of 10 out of the total 100 environments, with a maximum duration per run of \SI{100}\second. All headings and starting positions are randomized after each generation. Agents are spawned in some part of the area so that a path exists towards the gas source. 

We assess training results for training with doping. Table~\ref{tab:evo_vars} shows the evolved parameters in comparison with the manually designed parameters. $r_{range}$ is set to \SI{10}\meter, creating random waypoints in a box of \SI{10}\meter~in size around the agent during `exploring'. This box is scaled by evolved parameter $r_{r}$ (Equation~\ref{eq:no_gas}). When generating new waypoints, the agent has learned to move strongly towards the swarm's best-seen position $\mathbf{s}_{j}$, away from its personal best-seen position $\mathbf{p}_{i,j}$, and towards its previously computed goal $\mathbf{g}_{i,j-1}$. We expect this to be best in the environments encountered in evolution, as in most cases only one optimal position (with the highest concentration) exists. Hence, it does not need its personal best position to avoid converging to local optima. $\omega$ adds `momentum' to the waypoints generated, increasing stability.

$d_{swarm}$ shows that attraction-repulsion swarming is engaged only when another agent is within \SI{0.782}\meter. This is substantially lower than the manually chosen \SI{1.5}\meter, which can be explained by a lower cost when agents stay close to each other after finding the source. It also makes it easier to pass through narrow passages, but could increase the collision risk when agents are close to each other for a long time. Furthermore, $d_{wp}$ is set to \SI{2.69}\meter~, which is much higher than our manual choice and the timer has been practically disabled ($t_{wp}$ = 51.979). Instead of using the timeout, the evolved version uses PSO to determine the desired direction to follow, until it has travelled far enough and generates a new waypoint. For obstacle avoidance we see that the manual parameters are more conservative than the evolved counterparts. Being less conservative allows the agents to get closer to the source quicker, at the cost of an increased collision risk. This is an interesting trade-off, balancing individual collision risks with more efficient gas source localization.


After training, the probability for each environment in each draw can be evaluated as a measure of difficulty. Figure~\ref{fig:doping_props} shows a histogram of all 100 probabilities along with some environments on the spectrum. Generally, more cluttered environments with more local minima are more difficult.



\subsection{Baseline Comparison in Simulation}
\label{sec:baseline_sim}
It is difficult to compare Sniffy Bug with a baseline algorithm, since it would need to seek a gas source in cluttered environments, avoiding obstacles with only four laser rangers and navigating without external positioning. To the best of our knowledge, such an algorithm does not yet exist. Hence, we replace the gas-seeking part of Sniffy Bug (PSO) by two other well-known strategies, leading to (i) Sniffy Bug with anemotaxis, and (ii) Sniffy Bug with chemotaxis. In the case of anemotaxis, inspired by~\cite{10.1088/1748-3190/abbd81}, waypoints are placed randomly when no gas is present, and upwind when gas is detected. Then, when the agent loses track of the plume, it generates random waypoints around the last point it has seen inside the plume, until it finds the plume again. Similarly, the chemotaxis baseline uses Sniffy Bug for avoidance of obstacles and other agents, but places waypoints in the direction of the gas concentration gradient if a gradient is present. The chemotaxis baseline determines the gradient by moving a small distance in $x$ and $y$ before computing a new waypoint, while the anemotaxis baseline receives ground-truth airflow data straight from AutoGDM. We expect that the baseline algorithms' sensor measurements - and hence performance - would transfer less well to real experiments. For chemotaxis we expect that estimating the gradient in presence of noise and drift of the sensor data will yield inaccurate measurements. Anemotaxis based on measuring the weak airflow in indoor environments will be even more challenging.

We evaluate the models in simulation on all 100 generated test environments, and randomize start position 10 times in each environment, for a total of 1,000 runs for each model. We record different performance metrics: 1) success rate, 2) average distance to source, and 3) average time to source. Success rate is defined as the fraction of runs during which at least one of the agents reaches within \SI{1.5}\m~from the source, whereas average time to source is the average time it takes an agent to reach within \SI{1.5}\m~from the source. For agents not reaching the source, \SI{100}\second~is taken as time to source.

\begin{table}[]
\begin{tabular}{l|l|l|l}
                  & Success Rate & \begin{tabular}[c]{@{}l@{}}Avg Distance\\  to Source {[}m{]}\end{tabular} & \begin{tabular}[c]{@{}l@{}}Avg time \\ to source {[}s{]}\end{tabular} \\ \hline
Manual params PSO  & 90 \%        & 3.06                                                                      & 51.58                                                                 \\

Manual params Anemotaxis  & 91  \%       & 4.12                                                                     & 60.56   \\
Manual params Chemotaxis & 80  \%     & 4.31                                                                     & 68.61 \\
Evolved PSO w/o Doping & 89 \%        & 2.95                                                                  & 44.33    \\
Evolved PSO with Doping & \textbf{92 \% }       & \textbf{2.75}                                                                      & \textbf{41.94}   \
\end{tabular}
\caption{Sniffy Bug evaluated on 100 randomly generated environments.}
\label{tab:sim_inference}
\vspace*{-0.6cm}
\end{table}

Table~\ref{tab:sim_inference} shows that Sniffy Bug with PSO outperforms anemotaxis and chemotaxis in average time and distance to source, and has a success rate similar to the anemotaxis baseline. Chemotaxis suffers from the gas gradient not always pointing in the direction of the source. Due to local optima, and a lack of observable gradient further away from the source, chemotaxis is not as effective as PSO or anemotaxis. Since we expect the sim2real gap to be much more severe for anemotaxis than for PSO, which only needs gas readings, we decide to proceed with PSO. 

\begin{figure}
\begin{minipage}[t]{0.49\linewidth}
    \centering
    \includegraphics[width=\linewidth]{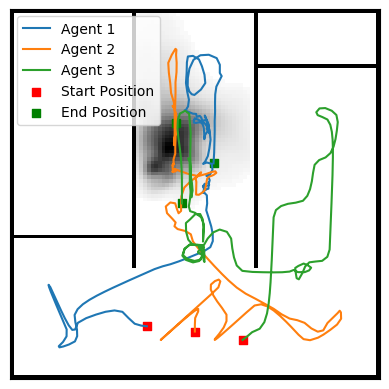}
    \caption{Sniffy Bug with manual parameters, successfully locating the source.}
    \label{fig:manual_trajectory}
\end{minipage}
\hfill
\begin{minipage}[t]{0.49\linewidth}
    \centering
    \includegraphics[width=\linewidth]{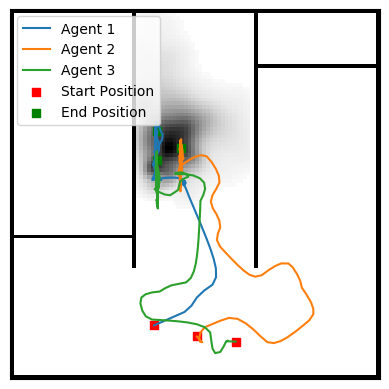}
    \caption{Sniffy Bug with parameters evolved using doping, finding the source quicker.}
    \label{fig:evolved_trajectory}
\end{minipage}
\vspace{-3mm}
\end{figure}
\subsection{Doping in Simulation}
\label{sec:inference_sim}

We now compare Sniffy Bug using PSO with manual parameters, parameters evolved without doping, and parameters evolved with doping. Table~\ref{tab:sim_inference} shows that the evolved parameters without doping find the source quicker, and with a smaller average distance to source, compared to the manual parameters. However, its success rate is slightly inferior to the manual parameters. This is caused by the hard instance problem~\cite{doping}: the parameters perform well on average but fail on the more difficult environments.

\begin{figure}
    \centering
    \includegraphics[width=\columnwidth]{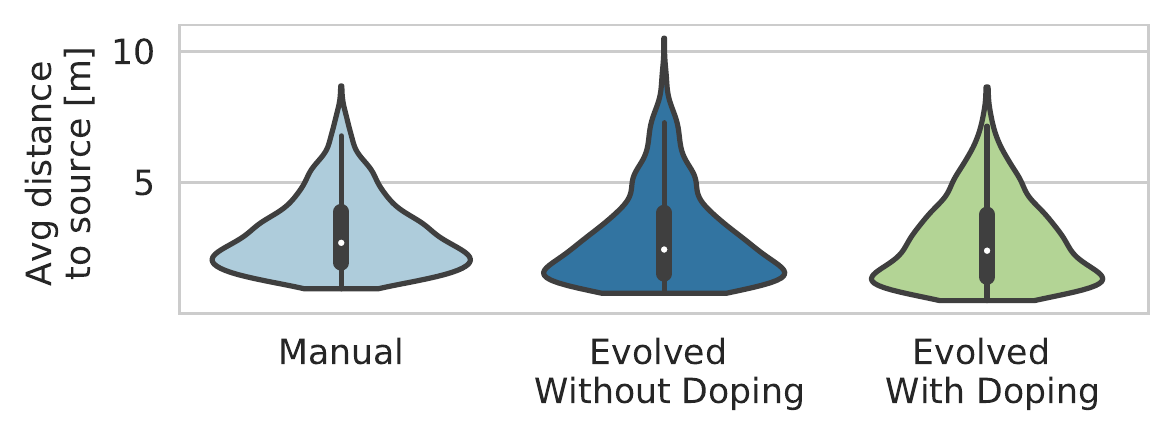}
    \vspace{-0.8cm}
    \caption{Three sets of parameters evaluated in simulation: 1) manually set parameters, 2) parameters evolved without doping, and 3) parameters evolved using doping~\cite{doping} to address the hard instance problem. }
    \label{fig:avg_distance_sim}
\end{figure}

Figures~\ref{fig:manual_trajectory} and \ref{fig:evolved_trajectory} show runs in simulation of the manual and evolved version with doping respectively, with the same initial conditions. The parameters evolved with doping outperform the other controllers in all metrics. Doping helps to tackle harder environments, improving success rate and thereby average distance to source and time to source. This effect is further exemplified by Figure~\ref{fig:avg_distance_sim}. The doping controller does not only result in a lower average distance to source, it also shows a smaller spread. Doping managed to reduce the size of the long tail that is present in the set of evolved parameters.

As we test on a finite set of environments, and the assumptions of normality may not hold, we use the empirical bootstrap~\cite{efron1979} to compare the means of the average distance to source. Using 1,000,000 bootstrap iterations, we find that only when comparing the manual parameters with evolved parameters with doping, the null hypothesis can be rejected, with $P=\num{2e-6}$, showing that parameters evolved with doping perform significantly better than manual parameters.


\subsection{Flight Tests}
\label{sec:flight_tests}
\begin{figure} 
    \centering
  \subfloat[Environment 1: the agents need to explore the `room' in the corner to smell the gas.\label{1a}]{%
      \includegraphics[width=0.49\linewidth]{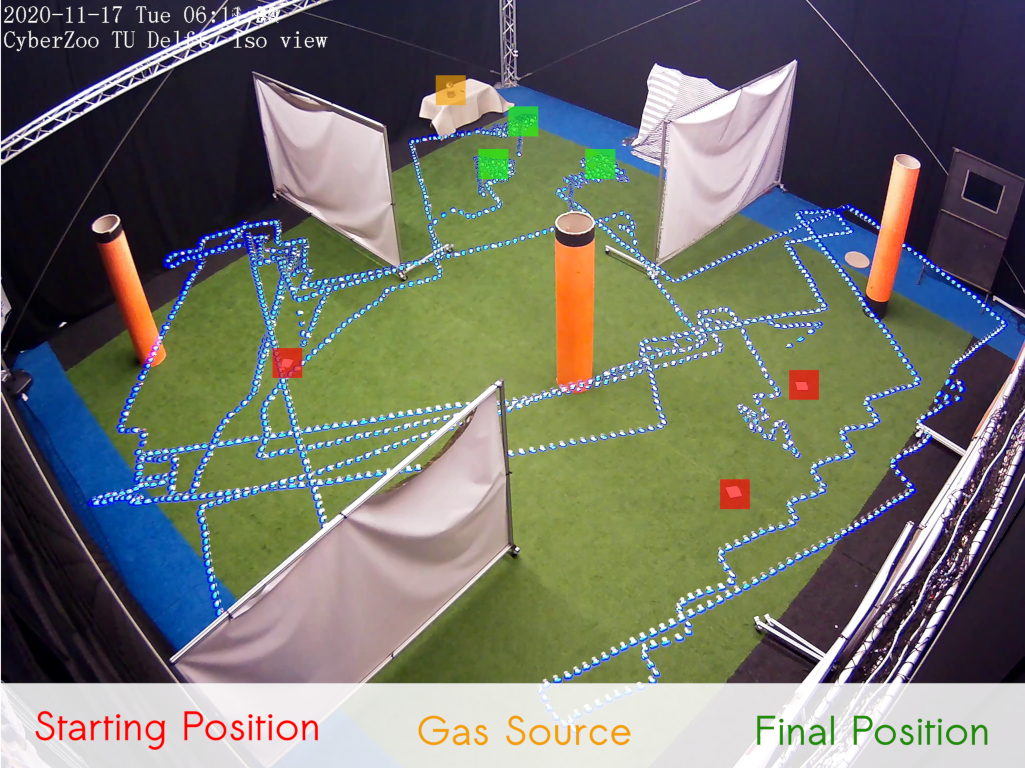}}
    \hfill
  \subfloat[Environment 2: the agents need to go around a long wall to locate the source.\label{1b}]{%
        \includegraphics[width=0.49\linewidth]{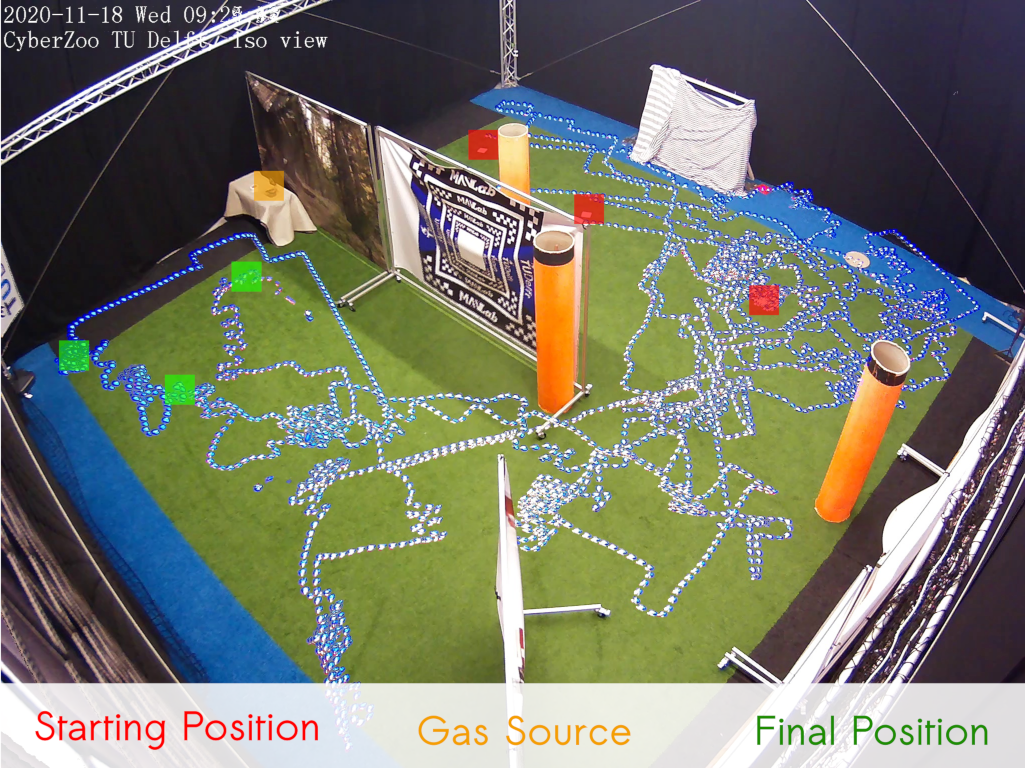}}
    \\
  \subfloat[Environment 3: an easier environment with some obstacles. \label{1c}]{%
        \includegraphics[width=0.49\linewidth]{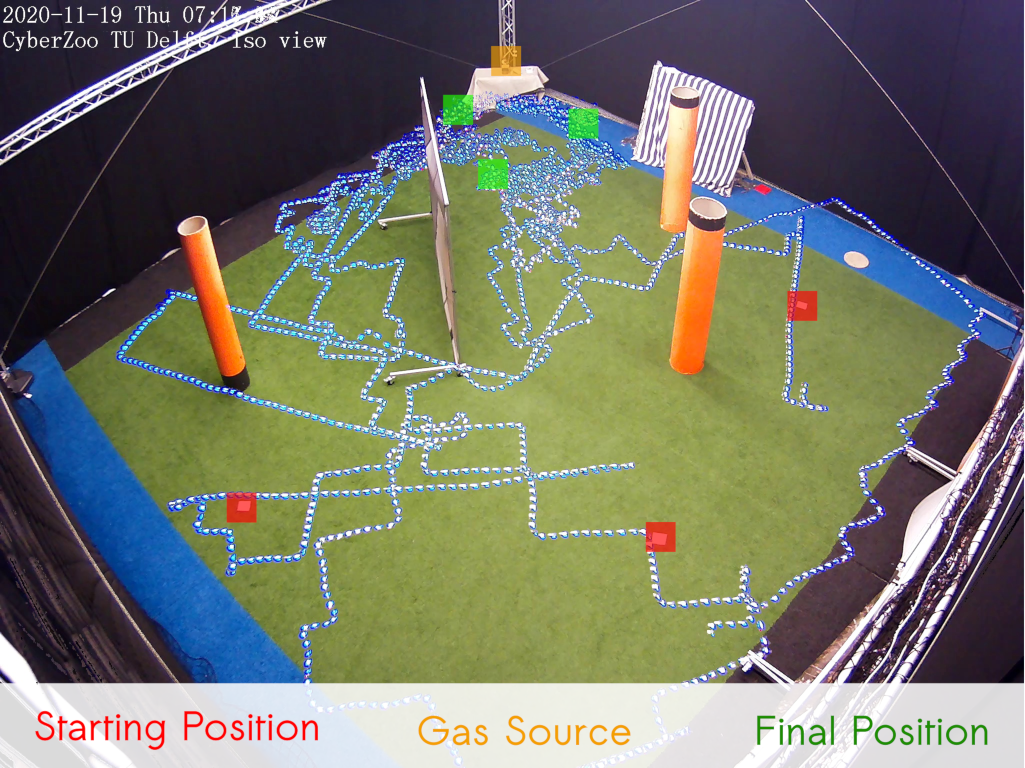}}
    \hfill
  \subfloat[Environment 4: an empty environment.\label{1d}]{%
        \includegraphics[width=0.49\linewidth]{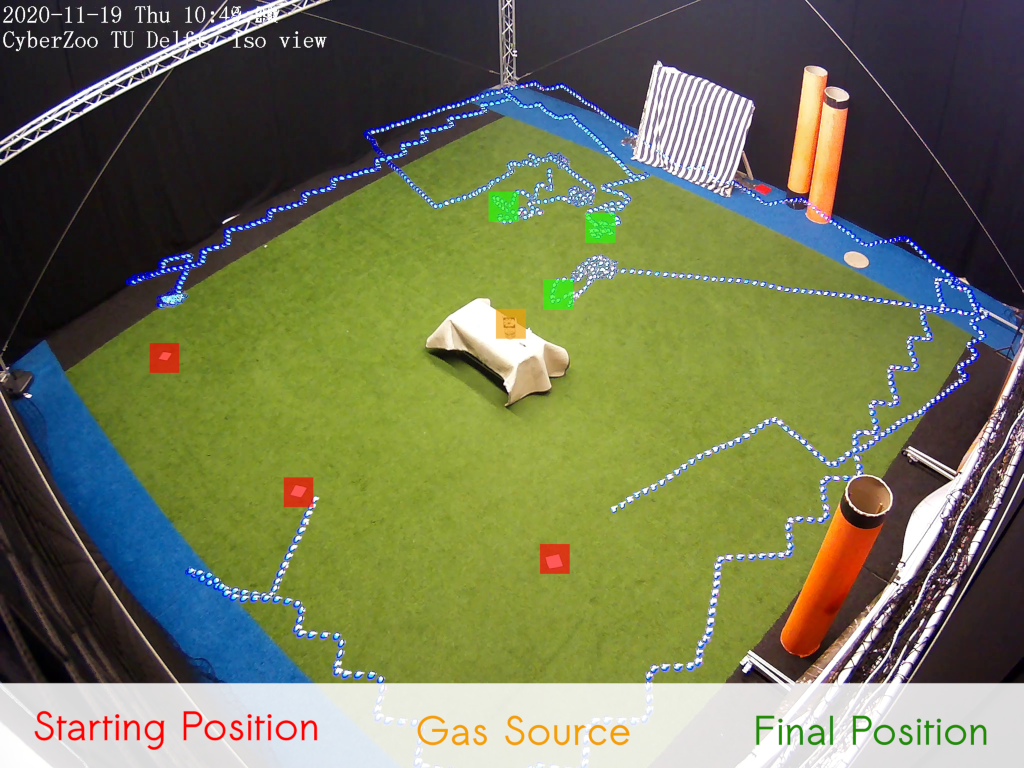}}
  \caption{Time-lapse images of real-world experiments in four distinct environment setups, 10 $\times$ 10 \si{\meter}~in size, seeking a real isopropyl alcohol source. The nano quadcopters' trajectories are visible due to their blue lights.}
  \label{fig:trajectories_flight}
  \vspace*{-0.2cm}
\end{figure}

\begin{figure}[htb!]
    \centering
    \includegraphics[height=\linewidth,angle=270]{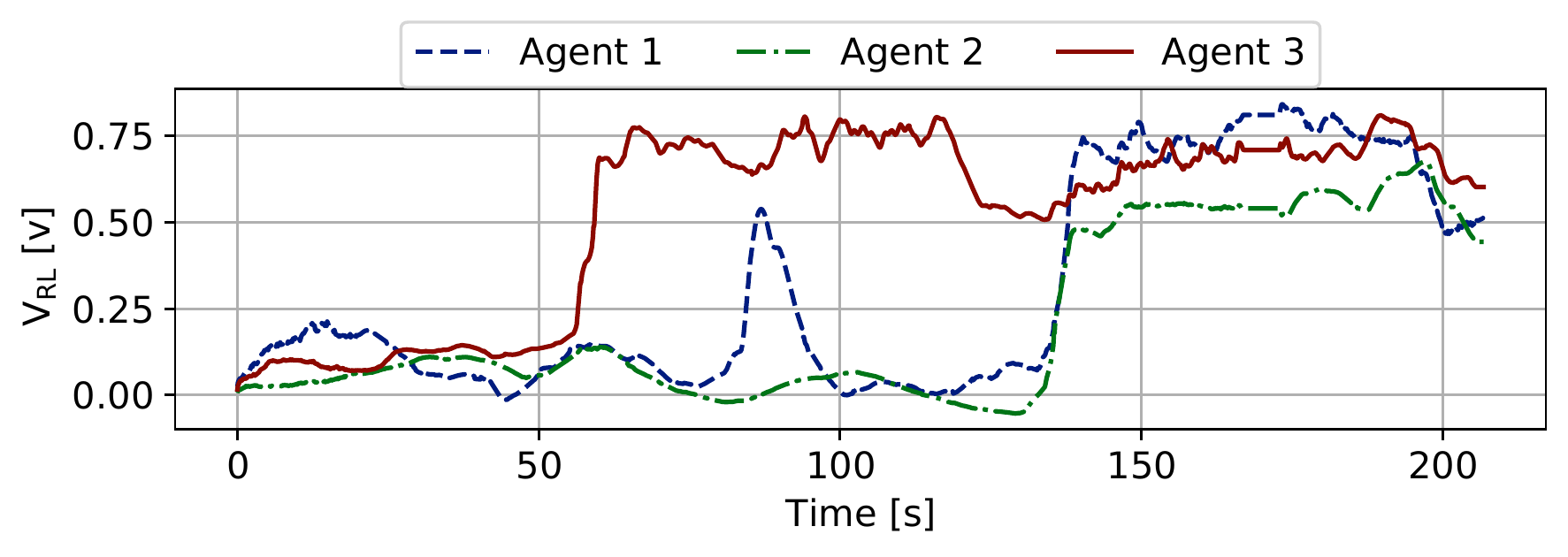}
    \caption{Evolved Sniffy Bug seeking an isopropyl alcohol gas source in environment 1. After agent 3 finds the source, all three agents quickly find their way to higher concentrations. }
    \label{fig:env_1_traces}
\end{figure}

Finally, we transfer the evolved solution to the real world, validating our methodology. We deploy our swarm in four different environments of 10 $\times$ 10 \si{\meter}~in size, as shown in Figure~\ref{fig:trajectories_flight}. We place a small can of isopropyl alcohol with a 5V computer fan within the room as the gas source. We compare manual parameters with the parameters evolved using doping, by comparing their recorded gas readings. Each set of parameters is evaluated three times for each environment, resulting in a total of 24 runs. A run is terminated when: 1) the swarm is stationary and close to the source, or 2) at least two members of the swarm crash. Each run lasts at least \SI{120}\second. Figure~\ref{fig:env_1_traces} corresponds to the run depicted in Figure~\ref{1a}.

\begin{figure} 
    \centering
  \subfloat[Environment 1.\label{2a}]{%
      \includegraphics[height=0.49\linewidth, angle=270]{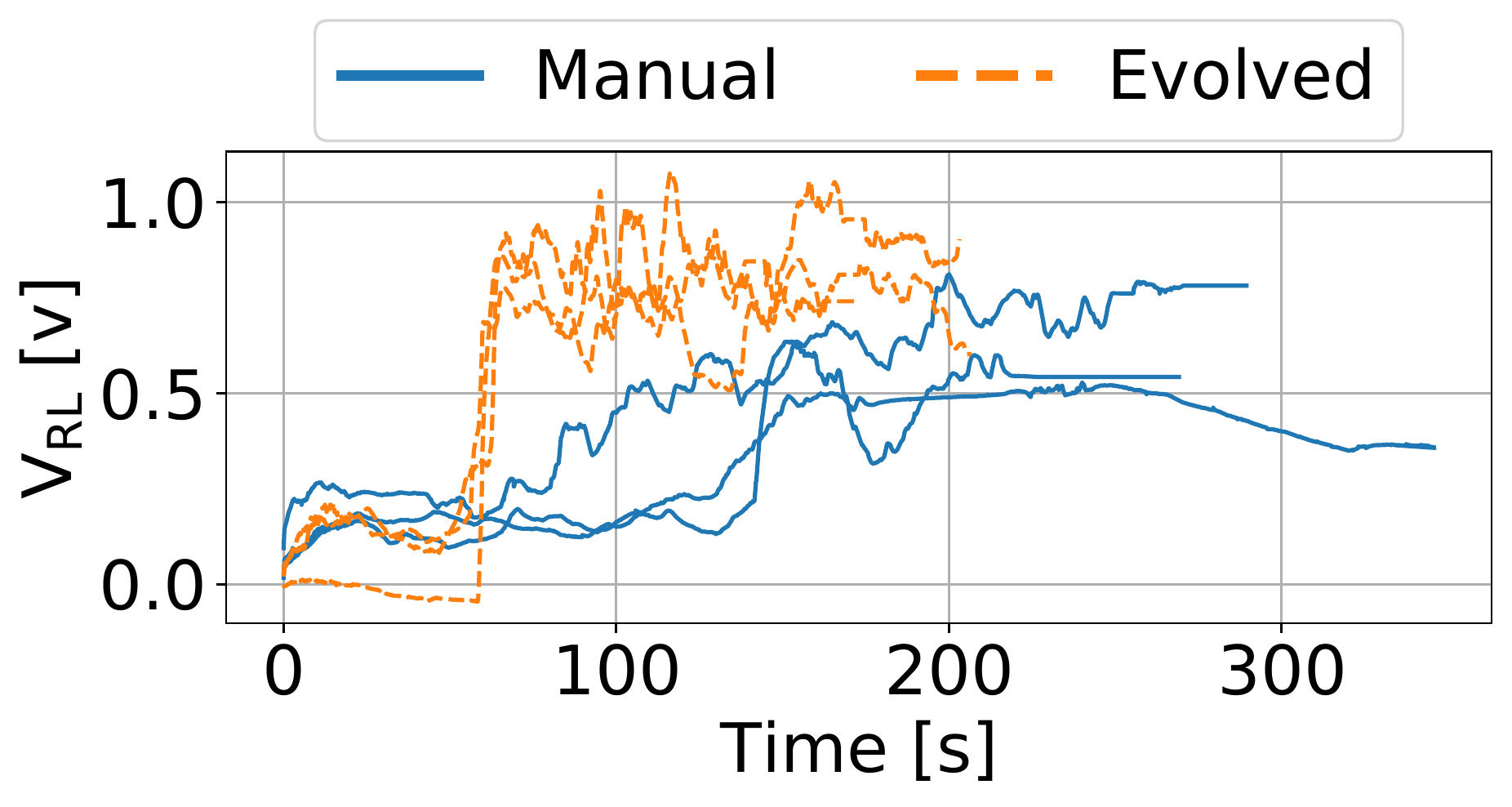}}
    \hfill
  \subfloat[Environment 2.\label{2b}]{%
        \includegraphics[height=0.49\linewidth, angle=270]{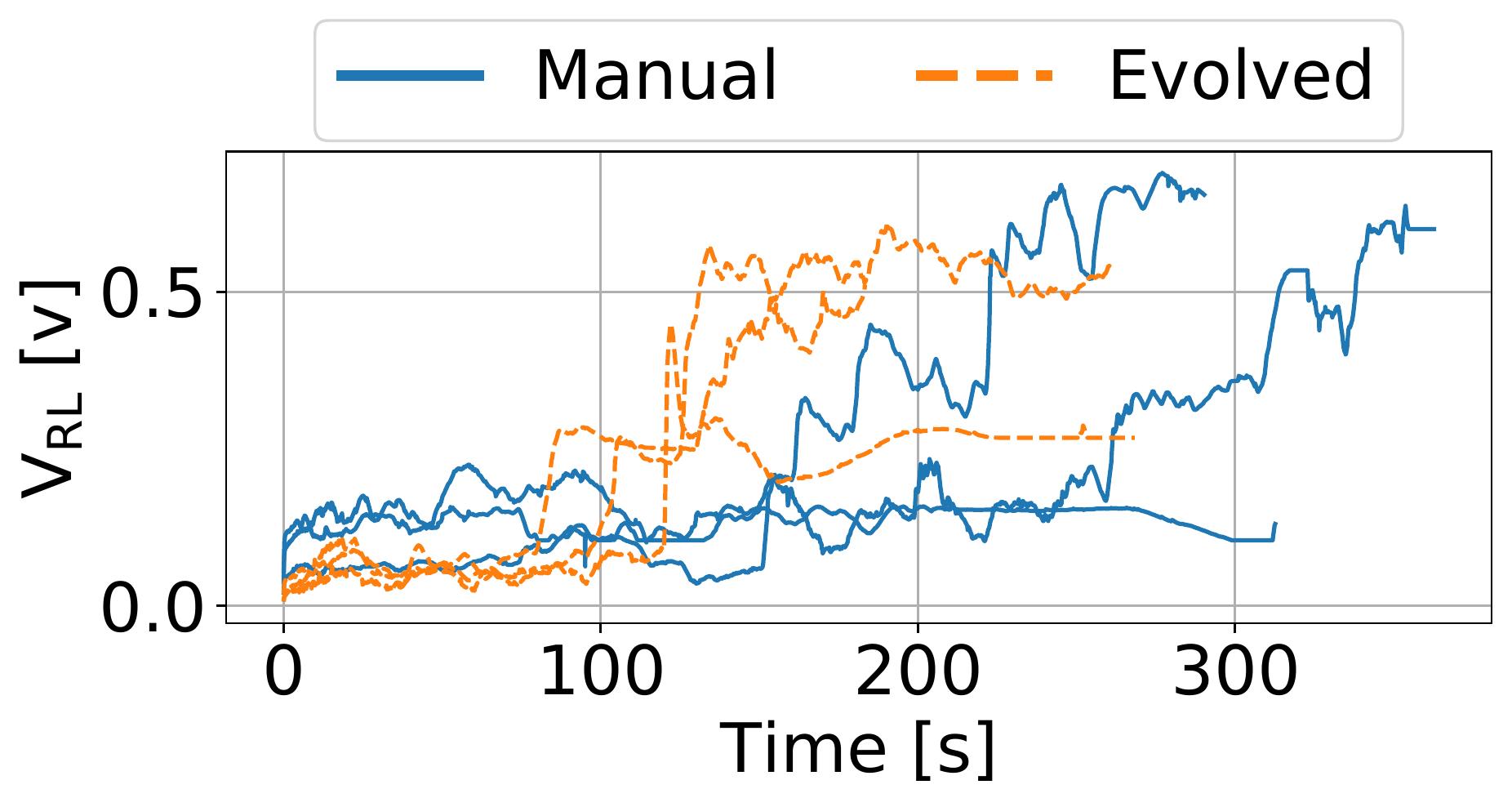}}
    \\
  \subfloat[Environment 3. \label{2c}]{%
        \includegraphics[height=0.49\linewidth, angle=270]{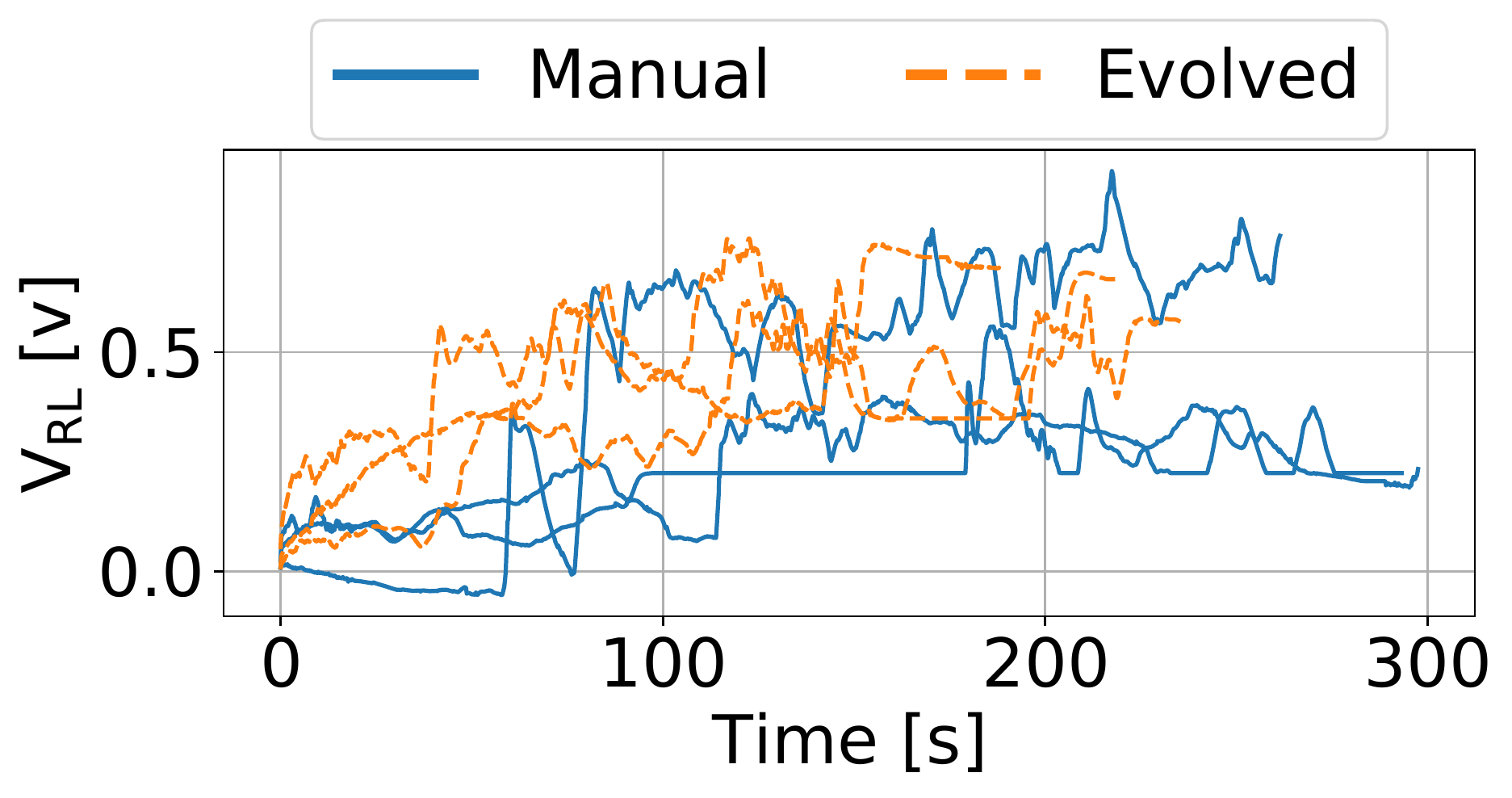}}
    \hfill
  \subfloat[Environment 4.\label{2d}]{%
        \includegraphics[height=0.49\linewidth, angle=270]{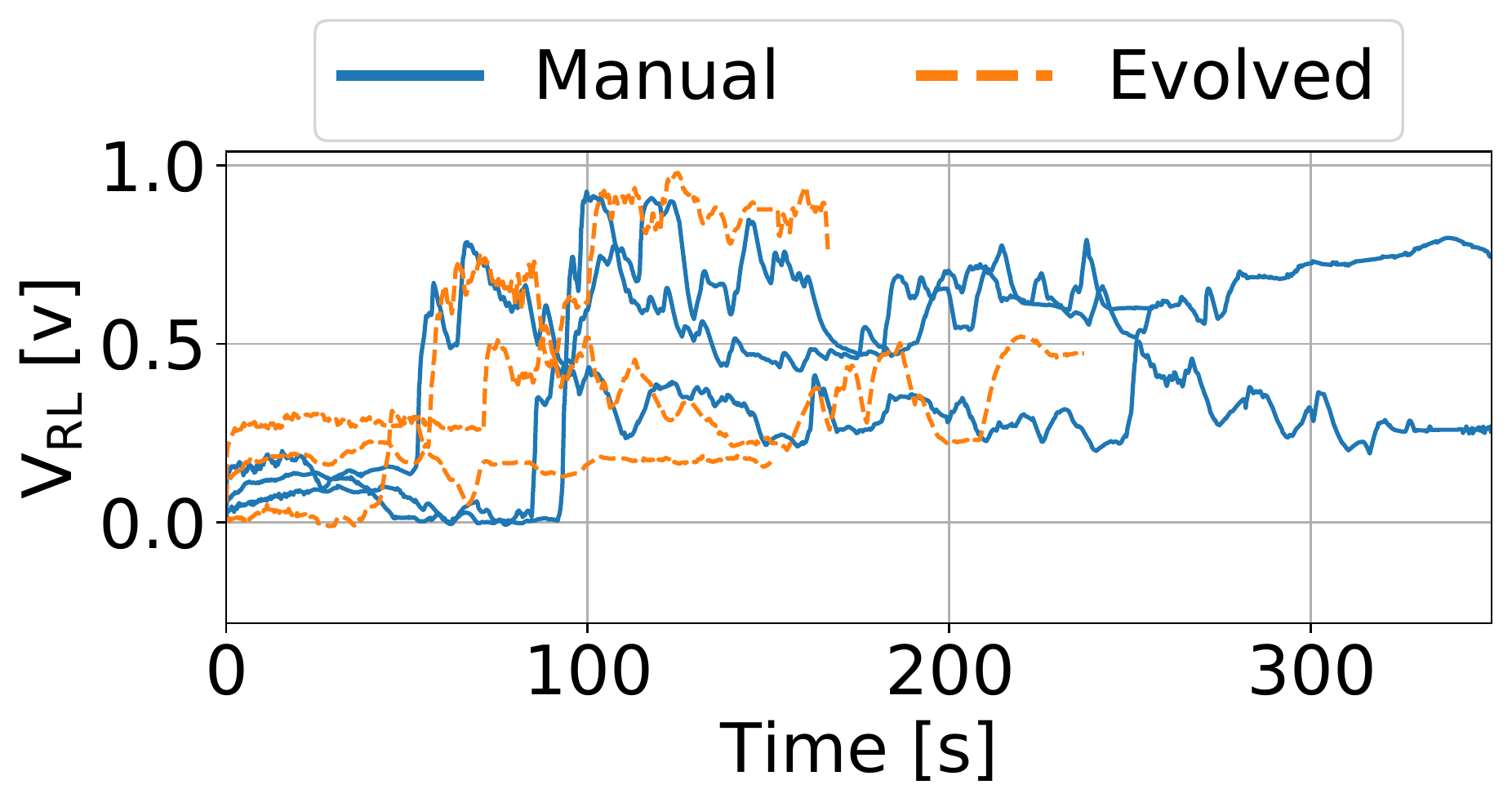}}
  \caption{Maximum recorded gas reading by the swarm, for each time step for each run. }
  \label{fig:max_all}
    \vspace*{-0.2cm}

\end{figure}

Figure~\ref{fig:max_all} shows the maximum recorded gas readings by the swarm, for each time step for each run. Especially for environment 1, it clearly shows the more efficient source seeking behavior of our evolved controller. Table~\ref{tab:all_traces} shows the average and maximum observed concentrations by the swarm, averaged over all three runs per environment. It shows that for environments with obstacles, our evolved controller outperforms the manual controller in average observed gas readings and average maximum observed gas readings. 

The evolved controller was able to reach the source within $\pm$\SI{2.0}\meter~in 11 out of 12 runs, with one failed run due to converging towards a local optimum in environment 4 (possibly due to sensor drift). The manual parameters failed once in environment 2 and once in environment 3. The manual parameters were less safe around obstacles, recording a total of 3 crashes in environments 1 and 2. The evolved parameters recorded only one obstacle crash in all runs (environment 2). We hypothesize that the more efficient, evolved GSL strategy reduces the time around dangerous obstacles. 

On the other hand, the evolved parameters recorded 2 drone crashes, both in environment 4, when the agents were really close to each other and the source for extended periods of time. The manual parameters result in more distance between agents, making it more robust against downwash and momentarily poor relative position estimates. This can be avoided in future work by a higher penalty for collisions during evolution, or classifying a run as a crash when agents are, for instance, \SI{0.8}\meter~away from each other instead of \SI{0.5}\meter.

\begin{table}[]
    \resizebox{.95\columnwidth}{!}{%
        \begin{tabular}{lll|ll}
              & \textbf{Manual}            &       & \textbf{Evolved}           &       \\
              & \multicolumn{1}{l|}{Avg $\pm$ std}   & Max $\pm$ std   & \multicolumn{1}{l|}{Avg $\pm$ std}   & Max $\pm$ std   \\ \hline
        Env 1 & \multicolumn{1}{l|}{0.250 $\pm$ 0.036} & 0.406 $\pm$ 0.049 & \multicolumn{1}{l|}{\textbf{0.330} $\pm$ 0.046} & 0.566 $\pm$ 0.069 \\ \hline
        Env 2 & \multicolumn{1}{l|}{0.162 $\pm$ 0.055} & 0.214 $\pm$ 0.070 & \multicolumn{1}{l|}{\textbf{0.165} $\pm$ 0.046} & 0.237 $\pm$ 0.063\\ \hline
        Env 3 & \multicolumn{1}{l|}{0.200 $\pm$  0.074} & 0.300 $\pm$  0.103& \multicolumn{1}{l|}{\textbf{0.258} $\pm$ 0.045} & 0.412 $\pm$ 0.029 \\ \hline
        Env 4 & \multicolumn{1}{l|}{\textbf{0.240} $\pm$ 0.123} & 0.398 $\pm$ 0.143 & \multicolumn{1}{l|}{0.176 $\pm$ 0.062} & 0.349 $\pm$ 0.151
        \end{tabular}
     }
 \caption{Average and maximum smelled concentration by the swarm, for manual and evolved parameters, averaged over 3 runs. }
 \vspace{-4mm}
\label{tab:all_traces}
\end{table}
 
The results show that AutoGDM can be used to evolve a controller that not only works in the real world in challenging conditions, but even outperforms manually chosen parameters. While GADEN~\cite{monroy_2017_gaden} and OpenFOAM~\cite{JASAK200989} are by themselves already validated packages, the results corroborate the validity of our simulation pipeline, from the randomization of the source position and boundary conditions to the simulated drones' particle motion model. 

\section{Conclusion}
\label{sec:conclusion}
We have introduced a novel bug algorithm, Sniffy Bug, leading to the first fully autonomous swarm of gas-seeking nano quadcopters. The parameters of the algorithm are evolved, outperforming a human-designed controller in all metrics in simulation and robot experiments. We evolve the parameters of the bug algorithm in simulation and successfully transfer the solution to a challenging real-world environment. We also contribute the first fully automated environment generation and gas dispersion modeling pipeline, AutoGDM, that allows for learning GSL in simulation in complex environments. 

In future work, our methodology may be extended to larger swarms of nano quadcopters, exploring buildings and seeking a gas source fully autonomously. PSO was designed to work in large optimization problems with many local optima, and is likely to extend to more complex configurations and to GSL in 3D. Finally, we hope that our approach can serve as inspiration for tackling also other complex tasks with swarms of resource-constrained nano quadcopters.



\bibliographystyle{IEEEtran}
\bibliography{IEEEabrv,references}

\end{document}